\documentclass[letterpaper]{article} 
\usepackage{aaai25}  
\usepackage{times}  
\usepackage{helvet}  
\usepackage{courier}  
\usepackage[hyphens]{url}  
\usepackage{graphicx} 
\usepackage{natbib}  
\usepackage{caption} 
\usepackage{algorithm}
\usepackage{algorithmic}
\usepackage{amssymb}
\usepackage{amsmath}
\usepackage{bm}
\usepackage{multirow}
\usepackage{booktabs}
\usepackage{arydshln}
\usepackage{pifont}
\newcommand{\xmark}{\ding{55}}  
\usepackage{mathrsfs}

\usepackage{newfloat}
\usepackage{listings}
\DeclareCaptionStyle{ruled}{labelfont=normalfont,labelsep=colon,strut=off} 
\floatstyle{ruled}
\newfloat{listing}{tb}{lst}{}
\floatname{listing}{Listing}
\title{RFL: Simplifying Chemical Structure Recognition with Ring-Free Language}
\author {
    Qikai Chang\textsuperscript{\rm 1},
    Mingjun Chen\textsuperscript{\rm 1},
    Changpeng Pi\textsuperscript{\rm 2},
    Pengfei Hu\textsuperscript{\rm 1},
    Zhenrong Zhang\textsuperscript{\rm 1},
    Jiefeng Ma\textsuperscript{\rm 1},
    Jun Du\textsuperscript{\rm 1}\thanks{Corresponding author.},
    Baocai Yin\textsuperscript{\rm 2},
    Jinshui Hu\textsuperscript{\rm 2}
}
\affiliations {
    \textsuperscript{\rm 1}NERC-SLIP, University of Science and Technology of China\\
    \textsuperscript{\rm 2}iFLYTEK Research\\
    \{qkchang,mingjunchen,pengfeihu,zzr666,jfma\}@mail.ustc.edu.cn, jundu@ustc.edu.cn, \{cppi,bcyin,jshu\}@iflytek.com
}

\usepackage{bibentry}

\begin{document}

\maketitle

\begin{abstract}
The primary objective of Optical Chemical Structure Recognition is to identify chemical structure images into corresponding markup sequences. However, the complex two-dimensional structures of molecules, particularly those with rings and multiple branches, present significant challenges for current end-to-end methods to learn one-dimensional markup directly. To overcome this limitation, we propose a novel Ring-Free Language (RFL), which utilizes a divide-and-conquer strategy to describe chemical structures in a hierarchical form. RFL allows complex molecular structures to be decomposed into multiple parts, ensuring both uniqueness and conciseness while enhancing readability. This approach significantly reduces the learning difficulty for recognition models. Leveraging RFL, we propose a universal Molecular Skeleton Decoder (MSD), which comprises a skeleton generation module that progressively predicts the molecular skeleton and individual rings, along with a branch classification module for predicting branch information. Experimental results demonstrate that the proposed RFL and MSD can be applied to various mainstream methods, achieving superior performance compared to state-of-the-art approaches in both printed and handwritten scenarios.
\end{abstract}

%
\begin{links}
    \link{Code}{https://github.com/JingMog/RFL-MSD}
\end{links}

\begin{figure}[tb]
\centering
\includegraphics[width=1.0\columnwidth]{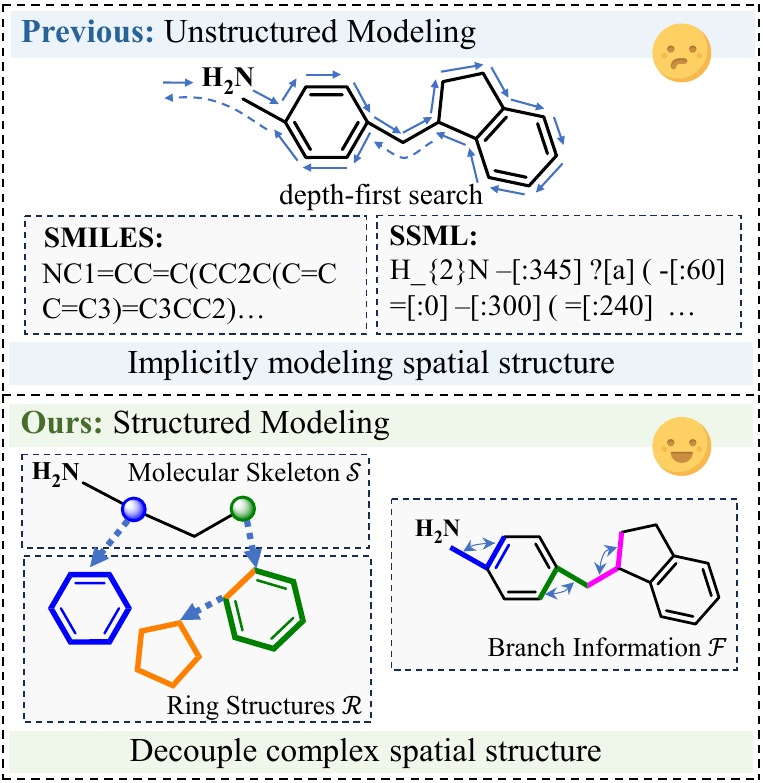}
\caption{Comparison of Ring-Free Language with previous modeling method. Previous methods use depth-first search to establish one-dimension markup, where spatial structures are implicitly expressed. Our RFL significantly simplifies complex spatial structures by decoupling them into molecular skeleton $\mathcal{S}$, ring structures $\mathcal{R}$ and branch information $\mathcal{F}$.}
\label{fig1}
\end{figure}

\section{Introduction}

Chemical structures, including molecules and formulas, are fundamental representations in chemistry. Optical chemical structure recognition (OCSR) \cite{rajan2020review} accurately converts handwritten and printed chemical structures into machine-readable formats across various media. This technology offers significant advantages in drug development, scientific research, the chemical industry, and education \cite{hu2023handwritten, oldenhof2024atom, chen2024icdar}. Unlike one-dimensional common text and tree-structured mathematical expressions \cite{zhang2020tree}, chemical structures are two-dimensional and graph-structured. Consequently, they contain more complex structures, such as rings and multiple branches. Therefore, models require not only robust capabilities to interpret the relationships between symbols and contexts but also advanced spatial analysis abilities.

Recent advances in computer vision have driven the development of bottom-up based OCSR techniques \cite{xu2022molminer, zhang2022abc, morin2023molgrapher, oldenhof2024atom, yao2024qe}. These approaches use multiple models and OCR engines to separately detect atoms and bonds, then reconstruct molecular structures. However, these methods are time-consuming, prone to error accumulation. Consequently, many recent studies on OCSR employ end-to-end algorithms \cite{weir2021chempix, xu2022swinocsr, xu2022molminer, hu2023handwritten, qian2023molscribe}, which are jointly optimized without additional modules. These algorithms mainly rely on image-to-markup frameworks, with markup typically in SMILES \cite{weininger1988smiles} or SSML \cite{hu2023handwritten}. In addition, many methods adopt motif-based modeling approaches to reduce sequence length \cite{yu2022molecular, zang2023hierarchical, jiang2023pharmacophoric}. These formats encapsulate complex two-dimensional molecular structures into one-dimensional markups. For instance, SSML uses depth-first traversal to generate strings, resulting in one-dimensional strings that represent molecular structures, as shown in Figure \ref{fig1}. Although this string-based representation is straightforward, there is no explicit design to guide the model in learning the spatial structure of the graph, leading to poor recognition performance in complex molecular structures with current end-to-end methods.

To address these issues, we propose a novel chemical formula modeling language named \textbf{R}ing-\textbf{F}ree \textbf{L}anguage (RFL). By equivalent conversion, RFL can decouple complex molecule structures and reducing the difficulty of the model's learning process. For a molecular structure $G$, it will be equivalently converted into a molecular skeleton $\mathcal{S}$, individual ring structures $\mathcal{R}$ and branch information $\mathcal{F}$, as shown in Figure \ref{fig1}. This approach effectively decouples and explicitly models the spatial structure of chemical molecules. Specifically, during graph traversal, individual rings are merged into a special atom, termed SuperAtom. Multi-ring structures are merged into a special bond, referred to as SuperBond, based on their adjacency until only individual rings remain, which are then merged into SuperAtom. The branch information $\mathcal{F}$ represents the mapping relationships between skeleton bonds and ring bonds, ensuring accurate reconstruction of the original molecular structure.

Based on Ring-Free Language, we propose a universal Molecular Skeleton Decoder (MSD) including a skeleton generation module and a branch classification module. The skeleton generation module is tasked with predicting $\mathcal{S}$ and $\mathcal{R}$, while branch classification module handles $\mathcal{F}$. Utilizing a divide-and-conquer strategy, the model first predicts the molecular skeleton, followed by ring structures, and finally reconstructs the molecular structure. Instead of parsing all complex molecular structures simultaneously, the model first parses the molecular skeleton and then individually predicts each ring structure, thereby reducing the errors.

We validate our method on the handwritten dataset EDU-CHEMC \cite{hu2023handwritten} and printed dataset Mini-CASIA-CSDB \cite{ding2022large}. We also select two representative mainstream methods as baselines to demonstrate the universality of our approach. Experiments show that our method significantly enhancing recognition ability for complex structures in both handwritten and printed scenarios.

Our primary contributions are as follows:

\begin{itemize}
    \item We propose \textbf{R}ing-\textbf{F}ree \textbf{L}anguage (RFL), which decouples complex molecular structures into molecular skeleton, ring structures, and branch information, thereby significantly simplifying the prediction process.
    \item Based on RFL, we propose a universal Molecular Skeleton Decoder (MSD) that can be applied to various mainstream approaches.
    \item Comprehensive experiments show that our method surpasses the state-of-the-art methods with different baselines on both printed and handwritten scenarios.
\end{itemize}

\section{Related Work}
\subsection{Bottom-Up Based OCSR}
During the early stages of OCSR development from 1990 to 2017, bottom-up based methods dominated the field. These approaches relied on hand-crafted rules to process images and detect all elements of chemical molecules, including chemical bonds, atomic groups, charges, and so on. Subsequently, they performed image vectorization to construct connection tables or graphs, thereby reconstructing the molecular structures \cite{filippov2009optical, fujiyoshi2011robust, smolov2011imago, ouyang2011chemink, sadawi2012chemical, bukhari2019chemical}. However, rule-based systems typically struggle with noisy images and are challenging and time-consuming to develop. 

With advancements in machine learning, numerous innovative techniques have emerged in OCSR. ChemGrapher \cite{oldenhof2020chemgrapher} and ABC-Net \cite{zhang2022abc} utilize Convolutional Neural Networks (CNNs) to predict molecular structures directly. MolGrapher \cite{morin2023molgrapher} employs a ResNet \cite{he2016deep} backbone to locate atom nodes within molecules and construct a Supergraph, which is then classified using a Graph Neural Network (GNN). Although graph-based methods perform well in complex printed scenarios, they often require additional OCR engines for support and are difficult to apply in complex handwritten scenarios. Consequently, many end-to-end methods have been developed to enhance recognition capabilities in various scenarios.

\subsection{Image Caption Based OCSR}
Most recent image caption based OCSR methods utilize an encoder-decoder end-to-end architecture. The encoder extracts features from the input image and passes them as input to the decoder. The decoder generates the final image caption, typically in SMILES \cite{weininger1988smiles}, InChI \cite{heller2015inchi} or SSML \cite{hu2023handwritten}.

MSE-DUDL \cite{staker2019molecular} employs a U-Net segmentation network \cite{ronneberger2015u} to extract and segment chemical structure diagrams from images. ChemPix \cite{weir2021chempix}, DECIMER \cite{rajan2020decimer}, and Img2Mol \cite{clevert2021img2mol} use CNN encoders and RNN decoders, such as LSTM, GRU, or standard RNN, to detect chemical structures. Recent works have also incorporated Transformer decoders, such as Transformer \cite{khokhlov2022image2smiles}, Swin Transformer \cite{xu2022swinocsr} and Vision Transformer \cite{sundaramoorthy2021end}. Additionally, some methods used sampler to improve performance \cite{yao2024swift}. 

However, these methods treat chemical molecular captions as simple strings without explicitly modeling them to enhance recognition performance, resulting in weak recognition performance in complex scenarios with current end-to-end solutions. Our new modeling language addresses this limitation by decoupling molecular structures through a divide-and-conquer strategy, thereby enhancing recognition capability for complex structures in end-to-end solutions.

\section{Ring-Free Language}
To decouple molecular structures, we propose a novel chemical structure modeling language called \textbf{R}ing-\textbf{F}ree \textbf{L}anguage (RFL). For a given molecular structure $G$, RFL converts it into an equivalent representation consisting of the molecular skeleton $\mathcal{S}$, individual ring structures $\mathcal{R}$ and branch information $\mathcal{F}$. The objective of the model is to separately predict $\mathcal{S, R}$ and $\mathcal{F}$, and ultimately reconstruct the original molecular structure. Each part of the RFL is separated by \texttt{[ea]} token.

The decoupling process adheres to specific principles: (1) Independent rings are merged into a SuperAtom, removed from $G$, and then appended into $\mathcal{R}$. (2) In cases involving adjacent multiple rings, the adjacency relationship between the rings, denoted as $\gamma$, is first resolved. The rings are then merged into a SuperBond in ascending order of $\gamma$, and subsequently appended into $\mathcal{R}$. The following sections will elaborate on the generation process of RFL, Splitting, and the graph restoration process, Restoring.

\subsection{Splitting}
The molecular structure is represented as a simple graph $G=(V, E)$, where $V$ denotes the set of vertices and $E$ denotes the set of edges. To decouple the ring structures, we first employ a modified depth-first search algorithm to detect all non-nested rings $\mathcal{R}$ in $G$. The set $\mathcal{R}$ is defined as follows:
\begin{equation}
    \mathcal{R} = \{C \in \mathcal{C} \mid \forall C' \in \mathcal{C}, (C' \subsetneq C \Rightarrow C' \not\subseteq C) \}
\end{equation}
where $\mathcal{C}$ represents all rings in $G$. According to the principles, we must analyze the adjacency relationships $\gamma$ between rings to determine the appropriate splitting strategy.

For a ring $r_i$, if $\gamma(r_i) = 0$, according to principle (1), $r_i$ will be merged into a SuperAtom and separated from graph $G$. The connection relationships $\mathcal{F}$ between $r_i$ and $G$ will be updated accordingly. $\mathcal{F}$ is defined as follows:
\begin{equation}
    \mathcal{F} = \{ (b_{i,j}, b_{u,v}) \mid b_{i,j} \in E_g \land b_{u,v} \in E_r \land j=u \}
    \label{connection}
\end{equation}
where $b_{i,j}$ represents the bond from $v_i$ to $v_j$, $E_g$ is the set of edges in $G$, and $E_r$ is the set of edges in $\mathcal{R}$. The direction of chemical bonds is determined during graph traversal, with a clockwise traversal adopted for rings. If $\gamma(r_i) > 0$, according to principle (2), $r_i$ will be merged into a SuperBond and separated from graph $G$. The SuperBond is one of the common bonds of $r_i$ and $r_j$. $\mathcal{F}$ will be updated accordingly. Ultimately, all $r_i$ are separated from $G$, resulting in a ring-free structure and yielding the molecular skeleton $\mathcal{S}$. 

\begin{figure}[tb]
\centering
\includegraphics[width=1.0\columnwidth]{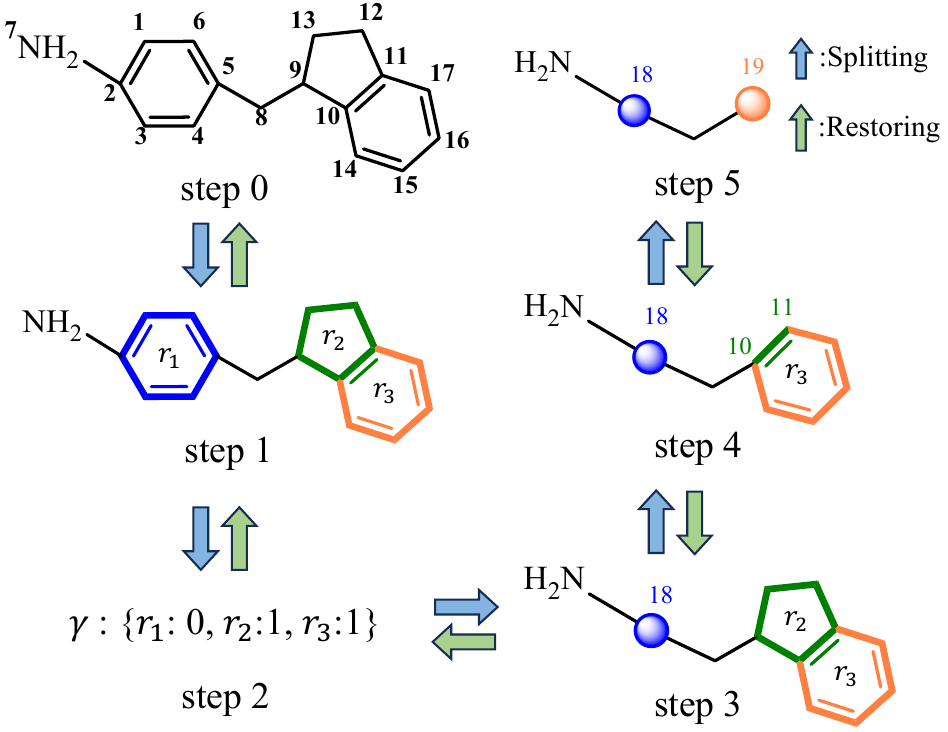}
\caption{The step-by-step decoupling process of Ring-Free Language, including the stages of Splitting and Restoring.}
\label{fig3}
\end{figure}

As an example, we will demonstrate how to apply the Splitting process to the molecular shown in Figure \ref{fig3}, starting from step 0. Step 1: A modified depth-first search algorithm is applied to obtain $\mathcal{R} = \{r_1, r_2, r_3\}$. Step 2: The adjacency relationships are analyzed, $\gamma = \{r_1 \colon 0, r_2 \colon 1, r_3 \colon 1\}$. In ascending order of $\gamma$, we merge the ring structures in sequence $(r_1, r_2, r_3)$. Step 3: For $r_1$, since $\gamma(r_1) = 0$, it is merged into SuperAtom $v_{18}$ and appended into $\mathcal{R}$. The branch information $\mathcal{F}$ is updated to $\mathcal{F}=\{(b_{7,2}, b_{2,1}), (b_{8,5}, b_{5,4})\}$. Step 4: For $r_2$, since $\gamma(r_2) = 1$, it is merged into SuperBond $b_{10, 11}$ and appended into $\mathcal{R}$. The branch information is updated to $\mathcal{F} \cup \{(b_{8,9}, b_{9,13})\}$. Step 5: For $r_3$, since $\gamma(r_3) = 0$, it is merged into SuperAtom $v_{19}$. Finally, the resulting sets of SuperAtom $A=\{v_{18}, v_{19} \}$, SuperBond $B=\{b_{10,11}\}$.

\subsection{Restoring}
The process of Restoring is the reverse of Splitting. In this procedure, individual ring structures $\mathcal{R}$ are sequentially restored into the molecular skeleton $\mathcal{S}$, replacing the SuperAtom and SuperBond. The branch information $\mathcal{F}$ is consumed until all rings are restored, at which point $\mathcal{F}=\varnothing$.

Take the molecular skeleton in Figure \ref{fig3} step 5 as an example. The set of SuperAtoms $A = \{ v_{18}, v_{19} \}$ and SuperBonds $B=\{b_{10,11}\}$. In the reverse order of Splitting, $r_3$ is first restored into $\mathcal{S}$ to replace SuperAtom $v_{19}$. At this stage $\mathcal{R}=\{r_1, r_2 \}$, $A = \{v_{18}\}$, $B=\{b_{10,11}\}$ and $\mathcal{F}=\{ (b_{7,2},b_{2,1}), (b_{8,5}, b_{5,4}), (b_{8,9}, b_{9,13}) \}$, as shown in step 4. Next, $r_2$ is restored into $\mathcal{S}$, replacing SuperBond $b_{10, 11}$. At this stage, $\mathcal{R}=\{ r_1 \}$, $A = \{ v_{18} \}$, $B=\varnothing$ and $\mathcal{F} = \{ (b_{7,2},b_{2,1}), (b_{8,5}, b_{5,4}) \}$, as shown in step 3. Finally, $r_1$ is restored into $\mathcal{S}$, replacing SuperAtom $v_{18}$. At this final stage, $\mathcal{R}=\varnothing$, $A=\varnothing$, $B=\varnothing$ and $\mathcal{F}=\varnothing$, indicating the end of the restoration process, as shown in step 1.

\section{Molecular Skeleton Decoder}
\begin{figure*}[tb]
\centering
\includegraphics[width=2.1\columnwidth]{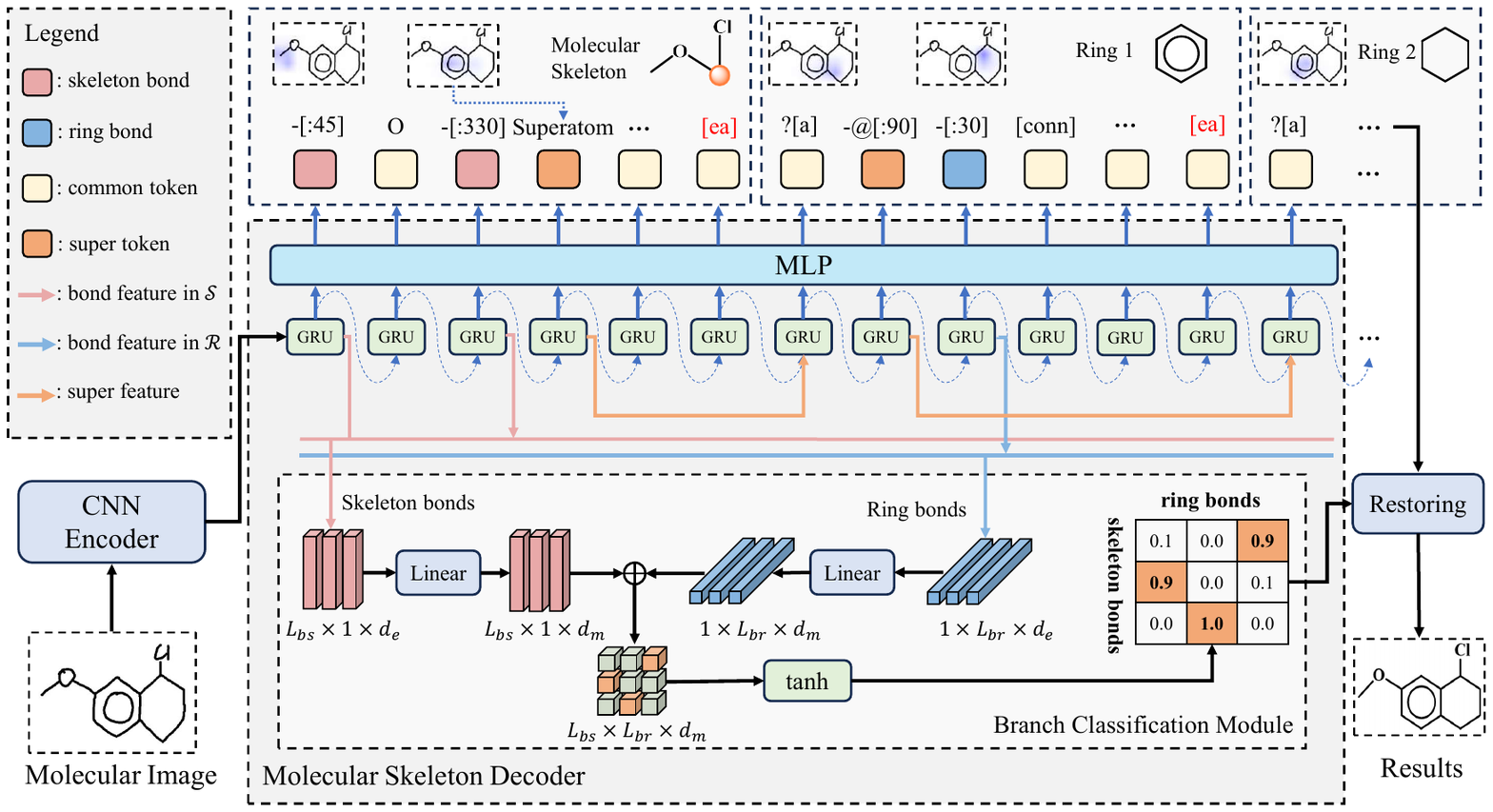}
\caption{The architecture of our method. First, the molecular image is input into the CNN Encoder to extract deep features. Subsequently, the molecular skeleton decoder autoregressively decodes the skeleton $\mathcal{R}$ and rings $\mathcal{S}$. During this process, the node features of skeleton and ring bonds are sent to the Branch Classification Module to derive branch information $\mathcal{F}$, which is used to restore the predicted molecular structure. The common token refers to tokens in RFL that are not specially processed.}
\label{architecture}
\end{figure*}

Based on the RFL, we propose a universal Molecular Skeleton Decoder (MSD) designed to leverage the advantages of RFL and enhance the model's recognition ability for complex molecules. The MSD consists of a skeleton generation module and a branch classification module. The overall architecture is shown in Figure \ref{architecture}. The skeleton generation module employs a hierarchical decoding approach to progressively predict molecular skeleton and rings. Subsequently, the node features of skeleton bonds and ring bonds are passed to the branch classification module to predict branch information, which represents the connection relationships between the molecular skeleton and individual rings. Finally, the complete molecular structure is restored.

\subsection{Skeleton Generation Module}
The skeleton generation module is responsible for predicting $\mathcal{S}$ and $\mathcal{R}$. It employs an encoder-decoder architecture, with the encoder based on DenseNet \cite{huang2017densely} and the decoder is a GRU \cite{chung2014empirical} model equipped with attention mechanisms \cite{zhang2018multi}. The DenseNet encoder extracts the feature map $\bm{F} \in \mathbb{R}^{C \times H \times W}$ from the input images, where $C$ represents the number of channels, $W$ and $H$ represent the width and height of the feature map. The decoder takes the feature map $\bm{F}$ as input and decodes each element in an autoregression manner. It adopts a hierarchical decoding approach: first detecting the molecular skeleton $\mathcal{S}$, then using the node feature vectors of SuperAtom and SuperBond as conditional information to continue predicting individual ring structures $\mathcal{R}$.

During the decoding process, when the model encounters ring structures, it does not fully parse them in one step. Instead, it first identifies and temporarily stores the node feature $\bm{f}^s$ of SuperAtom and SuperBond, then gradually decodes the ring structure only after the entire molecular skeleton has been fully decoded. With the node feature $\bm{f}^s$ as conditional information, the decoder continues decoding the ring structures once the molecular skeleton decoding is complete. Each node feature of the SuperAtom and SuperBond corresponds to a ring structure that needs to be decoded.
\begin{equation}
    \bm{f}^s_t = \operatorname{Concat}(\bm{E} \bm{y}_{t}, \bm{s}_{t}, \bm{c}_{t} )
\end{equation}
where $\bm{E} \in \mathbb{R}^{d_e \times V}$ represents word embedding matrix, $\bm{y}_{t} \in \mathbb{R}^{V}$ denotes the one-hot output vector at time $t$, $\bm{s}_t \in \mathbb{R}^{d_s}$ is the output state at time $t$, $\bm{c}_t \in \mathbb{R}^{d_c}$ is the context vector, and $V$ is the vocabulary size. $\bm{c}_t$ is calculated as follows:
\begin{align}
    \bm{c}_{t} = \sum_{i=1}^{H \times W} \bm{\alpha}_{ti} \bm{x}_i \\
    \bm{\alpha}_{ti} = \frac{\bm{e}_{ti}}{\sum_{k=1}^{H\times W} \bm{e}_{tk}}
\end{align}
\begin{equation}
\begin{aligned}[t]
    \bm{e}_{ti} = \bm{w}^T \tanh( & \bm{W}_x \bm{x}_i + \bm{W}_y \bm{E} \bm{y}_{t-1} + \bm{W}_s \bm{s}_{t} \\
    & + \bm{W}_a \bm{a}_i + \bm{W}_f \bm{f}^s_u)
\end{aligned}
\label{energy}
\end{equation}
where $\bm{w}, \bm{W}_x, \bm{W}_y, \bm{W}_s, \bm{W}_a, \bm{W}_f$ are trainable weights. The term $\bm{e}_{ti}$ represents the energy of $\bm{x}_i$ at time step $t$. The current attention weight $\bm{\alpha}_{ti}$ is obtained by applying the softmax function to $\bm{e}_{ti}$. The coverage vector $\bm{a}_i$, used to address the lack of coverage with historical attention weight information \cite{tu2016modeling}, is calculated as:
\begin{align}
    \bm{a}_i = \sum_{k=1}^{t-1} \bm{\alpha}_{tk}
\end{align}
To achieve hierarchical decoding, the node feature $\bm{f}^s_u$ is input as an additional condition into the decoder to calculate the energy $\bm{e}_{ti}$ at the current time step in Eq. (\ref{energy}), ensuring that the model can continue decoding individual ring structures from the time step $u$ corresponding to $\bm{f}_u^s$. When the model predicts \texttt{[ea]}, it switches the prediction phase, at which point $\bm{f}^s$ is a non-zero vector. The entire decoding phase ends when there are no SuperAtoms or SuperBonds left to decode and the end token \texttt{[END]} is decoded.

At every time step, the decoded token is:
\begin{align}
    P(\bm{y}_t | \bm{y}_{\leq t-1}, \bm{x}) = g(\bm{W}_o h(\bm{W}_c \bm{c}_t + \bm{W}_y \bm{E} \bm{y}_{t-1} + \bm{W}_s \bm{s}_{t}))
\end{align}
where $g$ denotes the softmax activation function, $h$ denotes the maxout activation function. A cross-entropy loss $\mathscr{L}_{ce}$ is used as training objective for skeleton generation module:
\begin{align}
    \mathscr{L}_{ce} = -\sum_{t=1}^{T} \log P(\bm{y}_t | \bm{y}_{\leq t-1}, \bm{x})
    \label{ce loss}
\end{align}

\subsection{Branch Classification Module}
The branch classification module is responsible for predicting $\mathcal{F}$, which represents the connection relationships between skeleton bonds and ring bonds, as described in Eq. (\ref{connection}). We construct a simple binary classifier to determine whether a connection exists between a skeleton bond $b_{s}$ and a ring bond $b_{r}$.
\begin{align}
    \bm{q}_{sr} = \tanh (\bm{W}_b \bm{f}_{bs}^s + \bm{W}_r \bm{f}_{br}^s)
\end{align}
where $\bm{W}_b, \bm{W}_r$ are trainable weights, $\bm{f}_{bs}$ is the node feature of skeleton bonds, $\bm{f}_{br}$ is the node feature of ring bonds. The output $\bm{q}_{sr} \in \mathbb{R}^{L_{bs} \times L_{br}}$ is the probability distribution of $\mathcal{F}$. The loss function of the branch classification module is formulated as follows:
\begin{align}
    \mathscr{L}_{cls} = -\sum_{s=1}^{L_{bs}} \sum_{r=1}^{L_{br}} \bm{y}_{sr} \log( \bm{q}_{sr} )
    \label{cls loss}
\end{align}
where $\bm{y}_{sr}$ is the one-hot vector of classification label.

The Branch Classification Module identifies the connected pairs from all candidate skeleton bonds and ring bonds. As illustrated in Figure \ref{fig1} below, many bonds on the rings are not connected to the skeleton. If all candidate bonds are fed into the branch classification module, the resulting classification matrix would be excessively sparse. To address this, a \texttt{[conn]} token is added after the connected ring bonds, ensuring that only the filtered ring bonds need to be classified, significantly reducing the number of candidate bonds. Essentially, the \texttt{[conn]} token divides the branch prediction task into two parts: predicting ring bonds and predicting skeleton bonds. The skeleton generation module is responsible for predicting ring bonds, while the branch classification module handles the prediction of skeleton bonds. This approach significantly reduces classification errors.

\section{Experiments}
\subsection{Implementation Details}
To ensure a fair comparison with the baseline methods DenseWAP \cite{zhang2018multi} and RCGD \cite{hu2023handwritten}, we utilize the same DenseNet encoder \cite{huang2017densely}, including three dense blocks. The growth rate and depth in each dense block are set to 24 and 32. The Molecular Skeleton Decoder (MSD) employs a GRU \cite{cho2014learning} with a hidden state dimension of 256. The embedding dimension is 256, and a dropout rate of 0.15 is applied.

The overall model is trained end-to-end. The training objective is to minimize the cross-entropy loss $\mathscr{L}_{ce}$ and the branch classification loss $\mathscr{L}_{cls}$. The objective function for optimization is defined as follows:
\begin{align}
    \mathcal{O} = \lambda_1 \mathscr{L}_{ce} + \lambda_2 \mathscr{L}_{cls}
\end{align}
In our experiments, we set $\lambda_1 = \lambda_2 = 1$. The Adam optimizer \cite{kingma2014adam} is used with an initial learning rate of $2 \times 10^{-4}$, and the parameters are set as $\beta_1=0.9, \beta_2=0.999, \varepsilon=10^{-8}$. The learning rate adjustment strategy employs MultiStepLR with a decay factor $\gamma=0.5$. All experiments are conducted on 4 NVIDIA Tesla V100 GPUs with 32GB RAM, using a batch size of 8 for the EDU-CHEMC dataset and 32 for the Mini-CASIA-CSDB dataset. The training epoch is set to 50, and the whole framework is implemented using PyTorch.

\subsection{Datasets}
We evaluate the performance of the proposed method on both handwritten dataset and printed dataset, as well as on our synthetic dataset.

\textbf{EDU-CHEMC} \cite{hu2023handwritten} contains 48,998 training samples and 2,992 testing samples of handwritten molecular structure images collected from various educational scenarios in the real world. The images are captured from various devices, including cameras, scanners, and screens. This dataset presents significant challenges due to its diverse writing styles and complex molecular structures.

\textbf{Mini-CASIA-CSDB} \cite{ding2022large} contains 89,023 training samples and 8,287 testing samples of printed molecular structure images collected from the chemical database ChEMBL \cite{gaulton2017chembl}. The images are rendered using RDKit \cite{landrum2013rdkit}. We preprocess the dataset using the same method as in \cite{hu2023handwritten}.

\subsection{Metrics}
For evaluation metrics, we use EM and Struct-EM to evaluate the performance of our model provided by CROCS-2024 \footnote{https://crocs-ifly-ustc.github.io/crocs/index.html}. EM measures the percentage of exact-match molecular images. Struct-EM measures the percentage of correctly identified structures in the predicted results, ignoring the non-chemical parts of the formulas. Struct-EM serves as an auxiliary evaluation metric.

\subsection{Comparison With State-of-the-Art}
To demonstrate the superiority of our method, we compare it with previous state-of-the-art (SOTA) methods. Table \ref{state-of-the-art table} presents the experimental results on the EDU-CHEMC dataset and Mini-CASIA-CSDB dataset. It can be observed that using DenseWAP \cite{zhang2018multi} as the baseline, MSD-DenseWAP achieves significant improvements on both EDU-CHEMC and Mini-CASIA-CSDB. Additionally, it surpasses the latest SOTA method on EDU-CHEMC by a notable margin of 2.06\%. This demonstrates the effectiveness of our method in enhancing the model's recognition ability. The relatively smaller improvement on the printed dataset Mini-CASIA-CSDB is mainly due to the lower proportion of complex molecules in this dataset.

\begin{table*}[t]
    \renewcommand{\arraystretch}{1.15}
    \centering
    \resizebox{1.0\textwidth}{!}{
    \begin{tabular}{lccccccc}
    \toprule
    \multirow{2}{*}{Methods} & \multirow{2}{*}{Markup} & \multirow{2}{*}{Params} & \multirow{2}{*}{Flops} & \multicolumn{2}{c}{EDU-CHEMC} & \multicolumn{2}{c}{Mini-CASIA-CSDB} \\ \cmidrule(r){5-6} \cmidrule(r){7-8}
     & & (M) & (G) & EM(\%) & Struct-EM(\%) & EM(\%) & Struct-EM(\%) \\ \midrule
    Imago \cite{smolov2011imago} \dag & SMILES & - & - & 0.00 & 0.00 & 38.80 & 38.88 \\
    WYGIWYS \cite{deng2016you} \ddag & SMILES & - & - & - & - & 78.55 & -* \\
    BTTR \cite{zhao2021handwritten} \ddag & SSML & 4.77 & 9.65 & 58.21 & 66.83 & 78.22 & -* \\
    ABM \cite{bian2022handwritten} \ddag & SSML & 22.45 & 19.39 & 58.78 & 67.24 & - & - \\
    CoMER \cite{zhao2022comer} \dag & SSML & 4.99 & 11.18 & 59.47 & 68.71 & 90.67 & 91.06 \\ \midrule
    DenseWAP (Baseline) \cite{zhang2018multi} & SSML & 15.39 & 18.98 & 61.35 & 69.68 & 92.09 & 92.47 \\ 
    \textbf{MSD-DenseWAP} (Ours) & RFL & 16.01 & 21.00 & \textbf{64.92} & \textbf{73.15} & \underline{94.10} & \underline{94.44} \\ \midrule
    RCGD (Baseline) \cite{hu2023handwritten} & SSML & 16.11 & 21.02 & 62.86 & 71.88 & 95.01 & 95.38 \\
    \textbf{MSD-RCGD} (Ours) & RFL & 16.74 & 23.04 & \textbf{65.39} & \textbf{73.26} & \textbf{95.23} & \textbf{95.58} \\ \bottomrule
    \end{tabular}
    }
    
    \caption{Comparison with state-of-the-art methods on handwritten dataset (EDU-CHEMC) and printed dataset (Mini-CASIA-CSDB) in \%. * indicates using SMILE as the learning target on Mini-CASIA-CSDB, resulting in the absence of Struct-EM. \dag denotes our reimplementation results, \ddag refers to results from \cite{hu2023handwritten}. MSD-DenseWAP and MSD-RCGD use DenseWAP and RCGD as baselines, respectively. All Params(M) and Flops(G) are measured with an input size of (1, 3, 200, 500).}
    \label{state-of-the-art table}
\end{table*}

\begin{figure*}[tb]
\centering
\includegraphics[width=2.1\columnwidth]{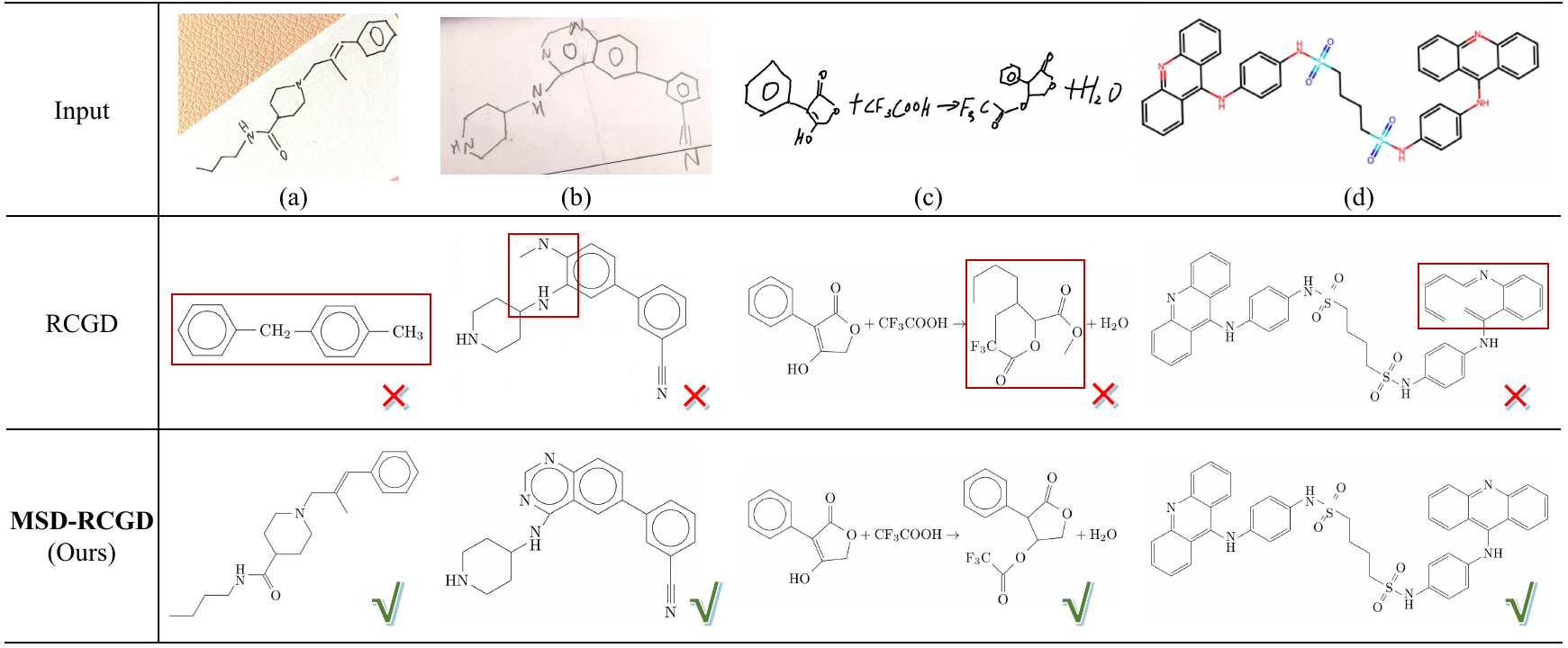}
\caption{A qualitative comparison of our proposed method with the SOTA method in the handwritten dataset EDU-CHEMC (a,b,c) and the printed dataset Mini-CASIA-CSDB (d). Compared to RCGD \cite{hu2023handwritten}, our approach robustly recognizes molecular structures in challenging complex ring structures.}
\label{case study}
\end{figure*}

To further validate the effectiveness and universality of our method, we select the latest SOTA method, RCGD \cite{hu2023handwritten}, as a baseline to construct MSD-RCGD. As shown in Table \ref{state-of-the-art table}, MSD-RCGD outperforms this baseline and achieves new SOTA results on both EDU-CHEMC and Mini-CASIA-CSDB. The qualitative comparison between MSD-RCGD and RCGD is illustrated in Figure \ref{case study}. These results demonstrate that our method can be applied to various mainstream end-to-end approaches, significantly enhancing recognition performance.

To explore the efficiency of our proposed method, we evaluate the parameters and FLOPs of MSD-DenseWAP and MSD-RCGD. The additional parameters and computational cost primarily arise from the branch classification module. As shown in Table \ref{state-of-the-art table}, the extra cost of our method is marginal. This demonstrates that our approach maintains efficiency while providing enhanced performance.

\subsection{Ablation Study}
To verify the effectiveness of our Ring-Free Language and Molecular Skeleton Decoder, we conduct ablation experiments through several designed systems in Table \ref{ablation table}.

\textbf{The effectiveness of MSD}. The Molecular Skeleton Decoder (MSD) is designed to fully leverage the advantages of RFL. To validate its effectiveness, we devised the systems T1 and T3, as shown in Table \ref{ablation table}. T3 utilizes MSD, whereas T1 does not, instead employing a standard string decoder. The experimental results show a significant improvement in the exact match rate when adopting MSD, demonstrating its effectiveness and necessity.

\textbf{The effectiveness of \texttt{[conn]}}. The branch prediction task includes two parts: predicting connected ring bonds and skeleton bonds. The \texttt{[conn]} token assigns these tasks to the Branch Classification Module and the Skeleton Generation Module, respectively. This approach decreases the classification errors. Therefore, we design systems T2 and T4, which incorporate the \texttt{[conn]} and achieve significant improvements compared to systems that do not use it. This is because incorrect branch information can prevent the correct reconstruction of the molecule, even if the skeleton and rings are predicted accurately.

\begin{table}[ht]
    \renewcommand{\arraystretch}{1.15}
    \centering
    \begin{tabular}{cccccc}
    \toprule
    System & MSD & \texttt{[conn]} & EM & Struct-EM \\ \midrule
    T1 & \xmark & \xmark & 38.70 & 49.45 \\ 
    T2 & \xmark & \checkmark & 44.02 & 55.77 \\ 
    T3 & \checkmark & \xmark & 52.76 & 58.58 \\ 
    T4 & \checkmark & \checkmark & 64.96 & 73.15 \\ 
    \bottomrule
    \end{tabular}
    \caption{Ablation study on the EDU-CHEMC dataset, with all systems based on MSD-DenseWAP.}
    \label{ablation table}
\end{table}

\subsection{Generalization Analysis}

\begin{figure}[th]
\centering
\includegraphics[width=1.0\columnwidth]{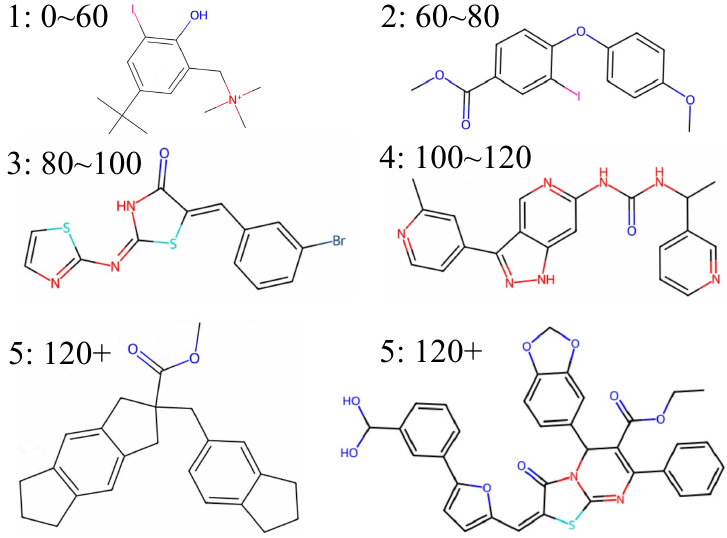}
\caption{Illustration of chemical molecules with increased structural complexity. The number in the top left indicates the structural complexity level of each molecule, with the corresponding complexity range.}
\label{complexity illustration}
\end{figure}

\begin{figure}[tb]
\centering
\includegraphics[width=1.0\columnwidth]{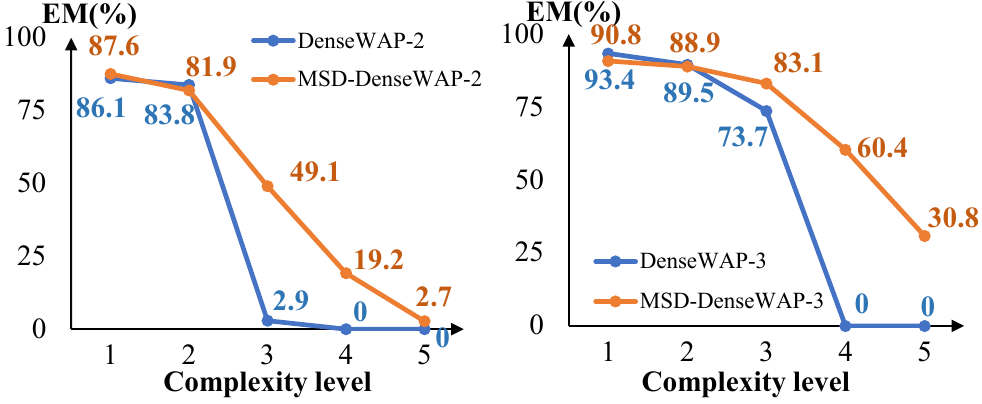}
\caption{Exact match rate (in \%) of DenseWAP and MSD-DenseWAP along test sets with different structural complexity. The left subplot is trained on complexity \{1,2\}, and the right subplot is trained on complexity \{1,2,3\}.}
\label{generalization figure}
\end{figure}

To prove that RFL and MSD can simplify molecular structure recognition and enhance generalization ability, we design experiments on molecule complexity. The previous approach \cite{hu2023handwritten} defined complexity as the sum of atoms and bonds. However, for complex multi-ring structures, recognition difficulty is much higher compared to ring-free structures with the same number of atoms and bonds. Therefore, we redefine structural complexity as:
\begin{align}
    Complexity = n_{atom} + n_{bond} + 12 \times n_{ring}
\end{align}
where $n_{atom}, n_{bond}, n_{ring}$ represent the number of atoms, bonds and rings, respectively. The coefficient 12 is chosen because the sum of atoms and bonds in the most common benzene rings is 12.

Due to the uneven distribution of structural complexities and the small proportion of complex samples in the Mini-CASIA-CSDB dataset, we created a new dataset from ChEMBL \cite{gaulton2017chembl} to assess model generalization. The dataset is divided into five levels based on structural complexity, with each level containing a similar number of samples, as shown in Figure \ref{complexity illustration}.

The experimental results are shown in Figure \ref{generalization figure}. DenseWAP-2 and MSD-DenseWAP-2 are trained with complexity levels \{1,2\}, while DenseWAP-3 and MSD-DenseWAP-3 use levels \{1,2,3\}. The test set includes all levels of structural complexity. It can be observed that when the structural complexity of chemical molecules in the test set exceeds the training set, DenseWAP-2 can hardly recognize them. However, MSD-DenseWAP-2 can still effectively recognize complex samples with previously unseen complexity. A similar trend is observed with MSD-DenseWAP-3, further confirming our conclusion.


\section{Conclusion and Future Work}
In this study, we propose a novel chemical structure modeling language \textbf{R}ing-\textbf{F}ree \textbf{L}anguage (RFL) to decouple complex ring structures in molecules. RFL decouples these structures into a molecular skeleton $\mathcal{S}$, individual rings $\mathcal{R}$, and branch information $\mathcal{F}$, thereby simplifying chemical structure prediction. Based on RFL, we propose a universal hierarchical Molecular Skeleton Decoder (MSD), which employs hierarchical decoding to progressively predict molecular skeleton $\mathcal{S}$ and individual rings $\mathcal{R}$, followed by restoring the structure using branch information $\mathcal{F}$. To the best of our knowledge, this is the first end-to-end solution that decouples and models chemical structures in a structured form. In the future, we plan to explore the extension of structured-based modeling to tasks like tables, flowcharts, and diagrams. Our goal is to extend the RFL-based divide-and-conquer philosophy to a broader range of applications and develop a unified modeling framework.


\nobibliography*
\bibliography{aaai25}

\end{document}